# Iterative collaborative routing among equivariant capsules for transformation-robust capsule networks

Sai Raam Venkataraman[1], S. Balasubramanian[2], R. Raghunatha Sarma[3]

[1]DMACS, SSSIHL,Prasanthi Nilayam,Puttaparthi,515134, Andhra Pradesh, India
[1]vsairaam@sssihl.edu.in,
[2]DMACS, SSSIHL, Prasanthi Nilayam, Puttaparthi, Anantpur, 515134, Andhra Pradesh, India
[2] sbalasubramanian@sssihl.edu.in
[3]DMACS, SSSIHL, Prasanthi Nilayam, Puttaparthi, Anantpur, 515134, Andhra Pradesh, India
[3]rraghunathasarma@sssihl.edu.in

*Abstract*
*Transformation-robustness is an important feature for machine learning models that perform image classification. Many methods aim to bestow this property to models by the use of data augmentation strategies, while more formal guarantees are obtained via the use of equivariant models. We recognise that compositional, or part-whole structure is also an important aspect of images that has to be considered for building transformation-robust models. Thus, we propose a capsule network model that is, at once, equivariant and compositionality-aware. Equivariance of our capsule network model comes from the use of equivariant convolutions in a carefully-chosen novel architecture. The awareness of compositionality comes from the use of our proposed novel, iterative, graph-based routing algorithm, termed Iterative collaborative routing (ICR). ICR, the core of our contribution, weights the predictions made for capsules based on an iteratively averaged score of the degree-centralities of its nearest neighbours. Experiments on transformed image classification on FashionMNIST, CIFAR-10, and CIFAR-100 show that our model that uses ICR outperforms convolutional and capsule baselines to achieve state-of-the-art performance.*

## 1. INTRODUCTION

What we see is a complex muddle of scenes, where objects and the scene themselves can occur in various confounded states such as being occluded or being in an unusual transformational state. While human perception is reliable for the most part, ongoing efforts continue to improve computer vision so that it is accurate and robust to confusions of input scenes.

Specifically, this paper deals with a part of this larger puzzle - namely, transformation-robust image classification. Image classification remains an ongoing challenge in the larger goal of scene perception. While one aspect of image classification aims at improving performance across varied objects and complex backgrounds, transformation-robust classification aims at preserving gains across transformations of the input scene. Transformation-robustness is also practically necessary as transformations of inputs are a major source of deviations of test scenes from the training scenes.

One means to bestow models with transformation-robustness is to use data-augmentation. Formal guarantees on the robustness of models, however, cannot be made for such techniques. It is for this reason that the equivariant models were proposed [1] [2]. For deep neural networks, these models preserve the representation of a scene across its transformations by having the hidden activations transform in the same manner as the input scene.

This leads to a mathematical notion of transformation-robustness where the hidden activations transform in the same way as the input. More formally, this is defined as the neural network preserving the group actions on an input. Equivariance, it must be noted, is a special property of the neural network, unlike the transformation-robustness offered by data augmentation.

How can equivariance be used for better transformation-robust learning? One means is to incorporate equivariant modules in a model that naturally uses its benefits for better learning of transformation-robust features. In this paper, we use equivariant modules to build a compositional, or part-whole aware deep neural network.

Our model is compositional in the sense that its intuition involves the detection of objects being based on the detection of its components, and them being in an appropriate configuration. Thus, the model takes into account part-whole relationships that are part of visual scenes. Learning these relationships is important to building transformation-robust models, as these relationships are invariant to transformations.

Specifically, our model is within the family of capsule networks that use vector activations, termed capsules, to model the poses of detected objects that are represented in their layers. Deeper capsules are built from shallower capsules by having the shallower capsules make predictions for the deeper capsules, and then combining these predictions in a manner that reflects the agreement in the predictions.

The method of combination can be done by techniques such as weighted-summation. The manner of obtaining these weights is related to the notion of agreement seen among the predictions. The weights used are referred to as routing-weights. The method of obtaining predictions involves the use of trainable neural networks with the capsules as input to them.

This process, termed routing-by-agreement, uses three assumptions that help explain its underlying intuition. First, that capsules represent objects that are detected by the corresponding layer and filter of a neural network. Second, that the predictions are seen as candidate poses for the object that deeper capsule represents. Third, that the routing-weights denote the extent of agreement among the candidate poses. Thus, the routing-weights can be seen to also denote the relationship between deeper and shallower capsules, along with being a notion of the extent to which the objects in the visual scene showcase a known compositional structure.

Under these assumptions, a capsule network ensures that objects in the hidden representation are formed based on whether their components are detected in an appropriate configuration.

Routing-by-agreement, when used in an equivariant setting leads to the following observation [3]. Equivariant routing assigns the same routing-weights to the same predictions, when seen for two inputs - one a transformed copy of the other. The only difference being that the weights and the predictions transform under the same transformation as the input.

Considering that capsules represent objects and routing-weights represent the part-whole relations amongst deeper and shallower capsules, equivariant routing may be seen as transformation-invariance in the detection of compositional relationships. This intuition is explained visually in Figure 1.

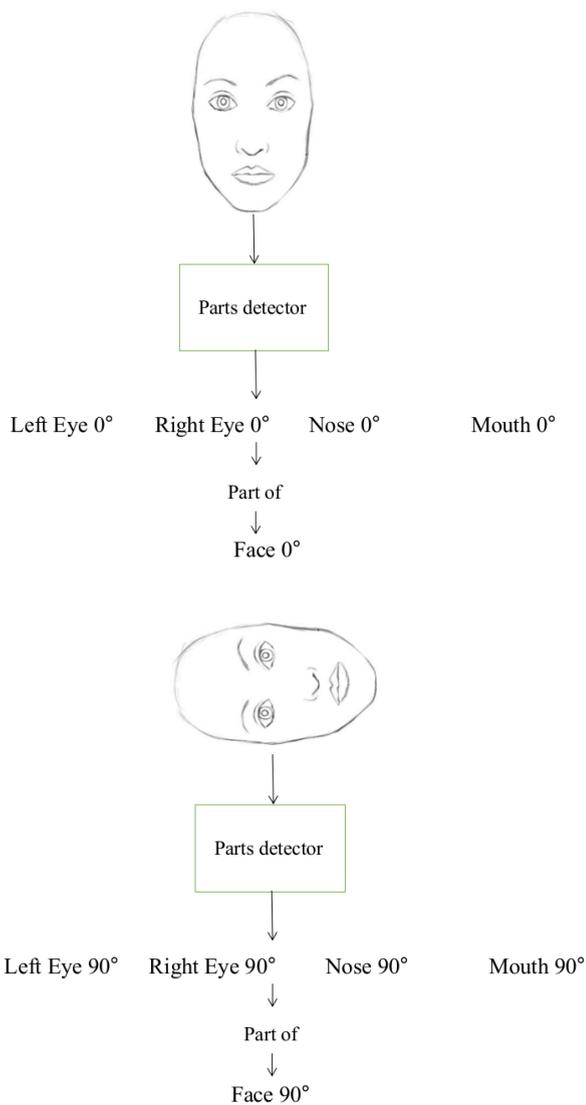

Fig. 1 The image above shows a part-whole structure for a face which is not rotated. The parts of the face are also unrotated. Consider the image below where the face is rotated by 90°. All the parts of the face are also rotated accordingly. The part-whole relationships are invariant to the transformation applied to the face. The face image is from [22].

Compositional models detect more fine-grained features that are useful for transformation-robust classification than non-compositional models do. Thus, we present a capsule network model that uses equivariant convolutions for this purpose.

In order for our model to learn compositional relationships, we present a novel, improved routing strategy that we term *Iterative collaborative routing* (ICR). ICR uses the iteratively averaged degree-centralities of closely-aligned predictions to build routing weights for predictions. In doing so, it uses neighbourhood information to present a novel and improved way of routing among capsules.

ICR also preserves equivariance - a property that makes it suitable for use in transformation-robust classification. This also allows it to share the intuition of transformation-invariant compositionality-detection that was presented in [3].

In order to demonstrate state-of-the-art performance on transformed image classification, we propose a strong architecture that could serve as a starting point for future work on capsule networks. Our architecture, with the use of ICR, achieves state-of-the-art classification accuracies on FashionMNIST, CIFAR-10, and CIFAR-100, outperforming several convolutional and capsule baselines.

Our work may be seen as a direction of showcasing the usefulness of the capsule inductive bias which has gone unexplored in recent days. Specifically, our results showcase that capsule networks, under appropriate conditions, can show good performance even in challenging conditions of train and test time.

In summary, our contributions are as follows:
- A novel equivariant routing procedure, termed Iterative collaborative routing (ICR) for capsule networks.
- A strong architecture for capsule networks that uses equivariant convolutions and ICR.
- State-of-the-art results on transformed image classification on FashionMNIST, CIFAR-10 and CIFAR-100.

The rest of this paper is organised as follows: Section 2 presents related work on equivariant neural networks and capsule networks; our proposed model is described in detail in Section 3; the experiments are presented in Section 4; the results of the experiments are discussed in Section 5; and the conclusion is presented in Section 6.

## 2. RELATED WORK

Equivariance is a well-studied property of neural networks, with multiple papers approaching the construction of equivariant networks across various groups, as well as building several network types.

The earliest of such works for modern deep neural networks is seen in [1], where the motivation and a basic formulation for equivariant convolutional networks was presented. This work models geometric transformations of the inputs using group actions, and argues for the preservation of these group actions in order to achieve transformation-robustness. The authors of this work extend the usual correlation of convolutional networks to be defined on general groups. This extension naturally brings about equivariance - the preservation of group actions, and results in transformation-robust classification. Two examples of this correlation were presented for two specific groups. These groups are formed from the composition of translations, orthogonal rotations, and reflections.

Several works, such as [2] and [4] extend the scope of equivariant convolutions to other , larger groups. For example, [2] presents convolutions for spherical transformations, while [4] presents a general framework for correlations equivariant to the euclidean group. Other works, such as [5] and [6] present applications of equivariant convolutions.

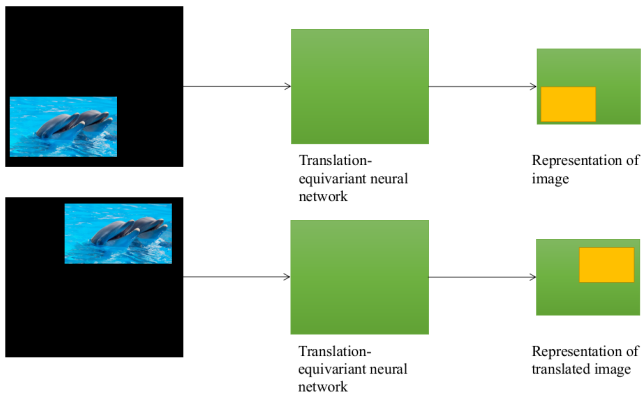

Fig. 2 The figure above shows the case of the detection of features (yellow) for an untransformed image. The figure below shows the same case for a translated image. A translation-equivariant network preserves the translations across its hidden representation. This is depicted by the translation of the yellow box that represents the detection of the features. Image is from [23].

The rationale behind the structure of capsule networks is based on learning the compositional or part-whole structure of objects in visual scenes. Early intuitions for this were laid out in [7]. The adaptation to modern deep neural networks was seen in [8] and [9], where backpropagation-based networks for obtaining predictions, and non-trainable routing algorithms were both used.

More recent capsule networks follow the basic structure of these models, but present variations in the manner of obtaining predictions and routing-weights. For example, [8] used fully-connected layers for obtaining predictions, and an iterative algorithm for routing-weights. [9] used fully-connected networks in a correlation-like manner along with an EM-based algorithm for routing-weights.

Other works, such as [10], used convolutional prediction-methods, while [11] used attention-based methods in routing. Most of these methods focused on improving classification performance without exploring the connections between equivariance and the compositional intuitions of capsule networks. Nonetheless, these models improved the structure of capsule models and provided starting points for high-performing equivariant models.

Equivariance in capsule networks is studied in works such as [3] and [12]. Both of these works, however, do not achieve state-of-the-art performance. This may be attributed, in part, to inefficient architectures used in the models. However, these models, and others [13], [14] show the benefits of introducing equivariance to capsule networks. [3], in particular, explores a connection between the preservation of detected compositionality and the equivariance of capsule networks.

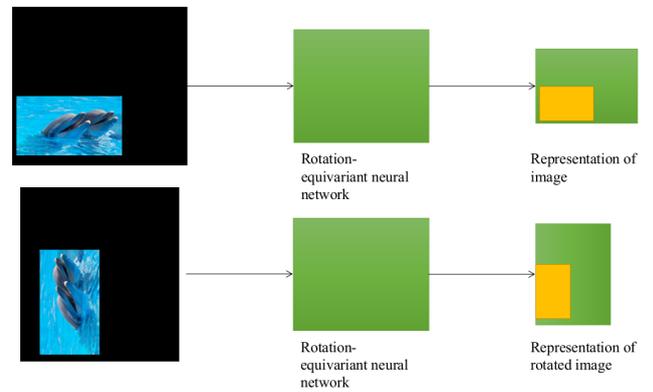

Fig. 3 The figure above shows the case of the detection of features (yellow) for an untransformed image. The figure below shows the same case for a rotated image. A translation-equivariant network preserves the rotations across its hidden representation. This is depicted by the rotation of the yellow box that represents the detection of the features. Image is from [23].

## 3. PROPOSED METHOD

In this section, we present a formal description of our capsule network model and the ICR algorithm. In order to do so, we first define preliminaries necessary to their formal definition. These are: the definition of general group-equivariant correlations and that of group-equivariance.

### 3.1 Preliminaries

We first provide the usual definition for correlations used in convolutional neural networks, following that with definitions for group-equivariant correlations and group-equivariance.

### 3.1.1 Correlation

Consider a function $f : Z^2 \to R^{d^l}$, that denotes the feature map output at the l-th layer of a neural network. Let $\Psi$ represent a set of $k^{l+1}$ filters of the l+1-th layer. Further, let $\Psi^i : Z^2 \to R^{k^l}$ denote the i-th filter of these. Further, let us use the notation $f_k$ to denote the k-th component of f, and $\Psi^i_k$ to denote the k-th component of $\Psi^i$. The correlation operator * is defined as:

$$[f * \Psi^i](x) = \sum_{y \in Z^2} \sum_{1 \le k \le d^l} f_k(y) \Psi^i_k(y - x).$$

This correlation operation has the property that it is equivariant to translations of the input. While we mathematically define equivariance in the next section of this paper, the actual consequences of this are simple. Figure 2 visually depicts equivariance to translations. Simply, this means that translations of the input are preserved to the output. The proof of this property is given in [1].

While the intuition of translation equivariance is simple, the authors of [1] argue that it is an important reason behind the success of convolutional neural networks. The gist of their argument is that natural symmetries in data, such as the invariance of label-information of objects when subjected to translations is captured in translation equivariance. This allows for a model that more closely resembles natural properties. Moreover, the correlation operation also allows for a detection of the same pattern shared across all translations, leading to lesser duplication of pattern detectors and, thereby, efficient neural networks.

While this correlation operation has shown itself to be a game changer of sorts for computer vision, and continues to be a major driver for deep learning, in terms of equivariance it does not completely satisfy observed natural properties.

Natural images display label-invariance to transformations that are more than translations. However, as was shown in [1], the correlation operation is not equivariant to them. In particular, it is not equivariant to rotations.

In order to have this property, [1] proposes a general definition of correlation that extends the usual definition. In this definition, the input and the filters are defined over a group of transformations as opposed to the grid-based definitions given earlier.

Formally, let G denote a group of transformations. Let $f: G \rightarrow R^{d^l}$, denote the feature map output at the l-th layer of a neural network. Let $\Psi$ represent a set of $k^{l+1}$ filters of the l+1-th layer. Further, let $\Psi^i: G \rightarrow R^{k^l}$ denote the i-th filter of these. Further, let us use the notation $f_k$ to denote the k-th component of f, and $\Psi^i_k$ to denote the k-th component of $\Psi^i$. The correlation operator * is defined as:

$$[f * \Psi^i](g) = \sum_{h \in G} \sum_{1 \leq k \leq d^l} f_k(h) \Psi^i_k(g^{-1}h).$$

Depending upon the group, a deep neural network using this correlation displays equivariance to transformations from that group. Figure 3 shows the case for rotation equivariance.

As can be seen, this correlation operation extends the usual correlation towards general transformations and aligns it with more general properties of observed symmetries.

The authors in [1] defined it for two groups: p4 - the group of translations composed with orthogonal rotations, and p4m - the group of translations composed with orthogonal rotations and reflections.

The use of this correlation on these groups has led to an increase in transformation-robustness. In particular, convolutional neural networks using equivariant correlations defined on p4 and p4m display greater robustness to rotations of the images. More general groups have also been studied, and further extensions of this correlation have been done in [2] and [4].

### 3.1.2 Equivariance

Figure 2 and Figure 3 convey the essence of group-equivariance: the preservation of transformations applied on the input of a function to its output. A formal description of this notion would necessitate a modelling of transformations and their effects on inputs and outputs.

In the equivariance literature, this modelling is done by using group actions. Group actions use intuition that transformations come from a group, and that the inputs and outputs come from vector spaces. Both of these assumptions are justified in that groups are a common way to model geometric transformations, while the inputs and outputs of neural networks are commonly seen as vectors in $R^n$.

We provide the formal definition of a group action first, and then the formal definition of equivariance.

**Group action:** Consider a group (G, .) and a vector space X. Then, a function $f: G \times X \rightarrow X$ is said to be a group action if the following are true:

- $f(e, x) = x, \forall x \in X$.
- $f(g_1, f(g_2, x)) = f(g_1 \cdot g_2, x), \forall g_1, g_2 \in G, \forall x \in X$.

In this work, as in most of the equivariant literature, we shall concern ourselves with linear group actions or group representations. Each element of a group can be associated with a matrix that comes from the group representation. Thus, given a group (G, .) and $g \in G$, a group representation for g can be written as a matrix $T_g$. This is the notation we shall use in this work. Equivariance, the preservation of transformations across the layers of a neural network, directly translates to preservation of the group representation.

**Group equivariance:** Consider a group (G, .) and a function $f: X \rightarrow Y$, where X and Y are vector spaces. Let T and T' be two group representations for G over X and Y, respectively. f is said to be equivariant with respect to T and T' if the following holds for all g: $f(T_g x) = T'_g(f(x)), \forall x \in X$.

For equivariance to correspond to the intuition described in Figure 2 and Figure 3, we have to use a specific kind of group representation. This group representation is given in [1]; we replicate its definition below.

$$[L_g f](x) = [f(g^{-1}x)].$$

The group-equivariant correlation defined previously satisfies the equivariance property
$$[[L_g f] * \Psi](x) = [L_g[f * \Psi]](x).$$

With the definition of group-equivariant correlations and group-equivariance as given, we now proceed to the definition of our proposed model.

### 3.2 Iterative collaborative routing among capsules

While equivariance is a mathematically guaranteed form of transformation-robustness, there are enhancements in model-structure that can add to it. Specifically, a model that utilises useful aspects of visual structure in a manner that incorporates equivariance naturally would be more robust to transformations than a model that does not utilise such intuitions.

One aspect of visual structure that remains invariant under symmetry transformations of the input is compositionality, or the part-whole structure of visual objects. Figure 4 demonstrates this transformation-invariance.

We consider the capsule network model for this purpose. It has been shown in [3] that the hidden representation between two capsule layers can be represented by a graph, where each prediction by a shallower capsule is treated as a vertex. Edges among vertices are constructed such that they exist only between predictions for the same deeper capsules, with an edge-weight given by a similarity between these predictions.

[3] also discusses how this *capsule-decomposition graph* is invariant (by isomorphism) under geometric transformations in an equivariant setting. Thus, in a capsule setting, it could be argued that a model which is able to learn relationships among the components of a visual object, and which is equivariant to those transformations can be thought of as preserving the detection of those relationships across transformations. Thus, equivariant capsule networks may be thought of as being transformation-invariant detectors of compositionality.

With this in mind, capsule networks can be considered as a viable candidate for transformation-robust models. We now describe how we construct an equivariant capsule network by describing how we construct each type of layer in our model.

### 3.2.1 Equivariant pre-capsule layers

Capsules are thought of as denoting objects, with shallower capsules being representative of components of deeper capsules. The role of routing algorithms is to build the deeper capsules in a manner that captures the agreement in the pose of the components.

The intuitions of capsules being representative of objects, and that of routing algorithms capturing relationships among those objects necessitates that the patterns dealt with in these layers correspond to a sufficiently high level complexity.

In other words, the first capsule layer that represents the simplest manner of objects must itself be input patterns that are complex enough to correspond to notions of parts and components.

With this in mind, and also keeping in mind the importance of equivariance, we propose to use a set of equivariant convolutional layers as initial pattern-detectors. This is based on the idea that stacking such layers leads to deeper layers capturing relatively complex features [15]. The use of such layers as a pre-processing of sorts for capsule layers has been done in capsule models such as DeepCaps [10] and STAR-CAPS [16].

In our model, we use 7 residual blocks that use equivariant convolutional layers. The equivariant correlations are defined on p4 - the group of transformations that can be thought of as compositions of translations and orthogonal rotations [1].

### 3.2.2 Equivariant primary capsule layer

The representation returned by the convolutional layers can be seen as a function that returns a scalar at each transformational state. These scalars can be thought of as the confidence of a pattern being present at a transformational state. These scalars must be converted to vectors that correspond to generalised poses of objects.

We propose to do this by using equivariant convolutions, one per dimension of the capsules. In order to have a uniform scale for these initial capsules, we follow the convolutions by a layernorm.

The capsules detected in this layer may be thought of as in the following manner. Each capsule represents an instance of a particular type of object detected at a transformational state. In other words, capsules are instances of archetypal capsule-types. This is ingrained into the model by the fact that each capsule can be seen as the result of the application of a set of specific equivariant filters to input scalars at specific transformational states. These filters, for each type, detect the important features for that type. Thus, the output of this layer can be thought of as a set of capsule-types, with each type denoting the capsules detected at various transformational states.

### 3.2.3 Capsule layers with iterative collaborative routing

In order to detect objects based on the learned compositionality present in them, and so as to ultimately classify objects, our model uses capsule layers that use equivariant correlations in conjunction with a novel equivariant routing algorithm. We describe the details of how these are used in the algorithm below. Before we describe the algorithm, however, we present the intuitions behind this.

First, given a set of input capsule-types, candidate capsules must be created for each deeper capsule. These candidate capsules can be thought of as poses of the object a deeper capsule represents, seen from the point-of-view of the shallower capsule. Our model uses the SOVNET-style of obtaining these predictions [3] in that each capsule-type is associated with a set of equivariant filters.

Each capsule-type obtains predictions for its capsules across transformational states by having capsules of the shallower capsule-types act as input to these filters. Unlike the SOVNET model, our model uses a single equivariant convolutional layer instead of a residual block.

The purpose of the computation of the routing-weights is to assign greater weights to predictions that are greatly similar to a large number of other predictions for the same deeper capsule. Moreover, an intuition for obtaining such weights is also that important predictions are, in turn, connected to other important predictions. This is akin to a notion of social importance, where individuals of influence are usually well-connected to other individuals of influence.

We model this idea by the following steps. We explain for a single capsule of a deeper capsule-type - the procedure is similar for other capsules of all capsule-types. First, we consider the predictions made by shallower capsules for a single capsule of a deeper capsule-type. We construct the affinity matrix of cosine similarities between all pairs of these predictions. From this, we construct the degree-centrality vector for the predictions. We then update the degree-centralities of the predictions by setting it to the mean of the degree-centralities of its nearest neighbours. The number of these neighbours is a hyper-parameter that is determined by experimental design. The nearest neighbours are determined by the affinity matrix.

This procedure is then repeated with the new degree-centralities using the same affinity matrix to obtain the nearest neighbours. The degree-centralities convey the notion of vertex-importance. By updating them based on the values of their nearest neighbours, ICR incorporates the notion of social importance that we discussed before.

After obtaining the weights for the predictions, the deeper capsule is formed by a weighted summation of the predictions using the softmaxed routing-weights. The algorithm describing this is given below. We provide the theorem showcasing the equivariance of ICR in a future section.

**Algorithm 1: Iterative collaborative routing**

**Input:** $\left\{ f_i^l \mid i \in \{0, ..., N_l - 1\}, f_i^l : G \to R^{d^l} \right\}$

**Output:** $\left\{ f_j^{l+1} \mid j \in \{0, ..., N_{l+1} - 1\}, f_j^{l+1} : G \to R^{d^{l+1}} \right\}$

**Hyper Parameters:** k, NUMITER

**Trainable functions:** $\left( \Psi_j^{l+1}, * \right)$, $0 \le j \le N_{l+1} - 1$, $*$ is the group equivariant correlation operator. $\Psi_j^{l+1}$ conveys a set of filters $\Psi_j^{l+1,p} : G \to R$, where $p \in \{0, ..., d^{l+1} - 1\}$.

1. $Pred_{ijp}^{l+1}(g) = \left( f_i^l * \Psi_j^{l+1,p} \right)$
2. $c_{0j}^{l+1}(g), ..., c_{N^l-1\,j}^{l+1}(g) = ICR\left( Pred_{0j}^{l+1}(g), ..., Pred_{N_{l+1}^l j}^{l+1}(g) \right)$
3. $f_j^{l+1}(g) = Squash\left( f_j^{l+1}(g) \right) = \frac{||f_j^{l+1}(g)||}{||1 + f_j^{l+1}(g)||^2}$

**Procedure:** $ICR\left( Pred_{0j}^{l+1}(g), ..., Pred_{0j}^{l+1}(g) \right)$

1. $A_{ik}^j(g) = \frac{Pred_{ij}^{l+1}(g) \cdot Pred_{kj}^{l+1}(g)}{||Pred_{ij}^{l+1}(g)|| \cdot ||Pred_{kj}^{l+1}(g)||}$
2. $DCen_i^j(g) = \Sigma_{k=0}^{N_l-1} A_{ik}^j(g)$
3. $KN_i^j(g) = NearestNeighbour(A, k, i, j)(g)$
4. For i = 1 to NUMITER: $DCen_i^j(g) = NearestNeighbourMean(KN, DCen_i^j(g), i, j)$
5. $c_{ij}^{l+1}(g) = \frac{\exp\left( DCen_i^j(g) \right)}{\Sigma_{i=0}^{N_l-1} \exp\left( DCen_i^j(g) \right)}$

**return** $c_{ij}^{l+1}(g)$

### 3.2.4. Projection to classification scores

The use of the capsule layers described before allows to create a hidden representation where intermediate object-shapes important to the classification of images can be learnt. This is extended to a final class-capsule layer, where the capsule-types of this layer represent the objects of the classes themselves.

The representation of this layer consists of vector capsules of various types. For classification, however, we require scalar scores that indicate the confidence of the presence of a particular kind of class-object. The conversion of vector capsules to scalar values was done in earlier works by considering a separate score for capsules, or by using the norm of the class capsules.

We instead use an equivariant convolutional layer that converts the number of dimensions to 1, while retaining the number of class capsules. The use of such a layer allows a learnable component in classification, thereby improving the accuracy of models.

### 3.3 Equivariance of ICR

In this section, we prove that the procedure ICR is equivariant when we use the group representation L.

**Theorem 1.** Algorithm 1 is equivariant.

**Proof:**
First, we see from the equivariance of * that $[[L_g f] * \Psi_j^{l+1}](x) = L_g[f * \Psi_j^{l+1}](x)$. Further, the product of two equivariant maps is also equivariant [1]. Moreover, the 2-norm of an equivariant function is also equivariant as it is the result of the post-composition of a pointwise non-linear map with an equivariant function [1]. Finally, the division of an equivariant map without another equivariant map results in an equivariant function [1]. Thus, the computation of $A_{ik}^j(g)$ and $DCen_i^j(g)$ at the steps 1 and 2 is equivariant.

From the equivariance of the computation of $A_{ik}^j(g)$, the computation of computation of $KN_i^j(g)$ is also equivariant. This is because the predictions as well as the cosine similarities at a transformational state g all get remapped to same transformational state $h^{-1}g$ upon application of a transformation by a transformation $h$.

From the equivariance of $KN_i^j(g)$ and that of computing the mean and the maximum function [1], the computation in the iterations are again equivariant. Finally, computation of softmax is again equivariant as it is a pointwise nonlinearity [1].

### 4. Experiments

### 4.1 Description of the datasets

We performed three sets of experiments that demonstrate the state-of-the-art transformation-robust classification performance of our model. Each set is performed on a separate, popular image classification dataset.

The images of the train dataset, for each of these datasets, are transformed by a random translation that has a maximum extent of upto 2 pixels in the horizontal and vertical directions, respectively. This is followed by a random rotation between -180° and 180°. Thus, the train dataset of the image classification datasets is significantly transformed so as to pose a challenge to models for learning in a noisy training setup.

Each of the test sets of the datasets we consider are transformed to yield 5 test datasets. Each of these test datasets corresponds to a level of geometric transformation-noise. The first of these is the untransformed test dataset itself. The second corresponds to a transformed version where images are first translated by a random translation with a maximum extent of upto 2 pixels, and then rotated by a random rotation between -30° and 30°. The other datasets are generated similarly, with their images being generated by transforming the original test images by a random translation and rotation. The extent for the translations is the same for all of these datasets, being upto 2 pixels in the vertical and horizontal directions. The rotational extent for the third dataset is between -60° and 60°, while it is between -90° and 90° for the fourth test dataset. The fifth, and hardest test dataset, has a rotational extent of -180° and 180°.

Thus, each classification dataset we consider is transformed so that the training data is transformed by high geometric noise, while the test dataset yields 5 versions that showcase various levels of geometric transformations. Good performance on the 5 test datasets shows that a model is able to learn in the presence of geometric transformations and generalise to situations that showcase transformations. This setup for testing this ability was formulated in [3] and [17].

We now present details about the datasets we have used, namely CIFAR10 [18], FashionMNIST [19], and CIFAR100 [18].

**CIFAR10:** The train dataset of CIFAR-10 consists of 50,000 32×32 images of 10 categories. Each category has 5000 images in the train dataset. The test dataset of CIFAR10 consists of 10,000 32×32 images of the 10 categories. Each category has 1000 images. CIFAR10 shows relatively diverse objects with variable backgrounds, making for a good challenge for transformation-robust classification.

**FashionMNIST:** The train dataset of FashionMNIST consists of 50,000 28×28 images of 10 clothing categories. The test set consists of 10,000 28×28 images of the 10 categories. Like CIFAR10, there are 5000 images of each category in the train dataset. Each category has 1000 images in the test dataset. FashionMNIST is relatively less of a challenge with respect to CIFAR10 as its images have a uniform background. However, it is still not trivial especially for transformed image classification.

**CIFAR-100:** The train dataset of CIFAR-100 consists of 50,000 32×32 images of 100 categories. Each category has 500 images in the train dataset. The test dataset of CIFAR100 consists of 10,000 32×32 images of the 100 categories. Each category has 100 images. CIFAR100 is harder to classify when compared to the other two datasets due to the larger number of categories and the diverse objects and backgrounds in the images.

**4.2 Description of the baselines**

In order to showcase the challenging nature of the task of transformation-image classification, we consider the following capsule baselines: CapsNet [8], EMCaps [9], GCaps [12], DeepCaps [10], SOVNET [3]. We also consider ResNet18, ResNet34 [19], and their group-equivariant versions defined over the p4m group: P4MResNet18 and P4MResNet34 [1]. The results for these models have been obtained from [3] and independent experiments.

**4.3 Implementation details**

In order to achieve the best possible results on image classification, certain hyperparameters were found empirically. Moreover, a change to Algorithm 1 is also implemented that causes a slight loss in exact equivariance, but improves empirical performance significantly.

We first describe the change in Algorithm 1 and also the reasoning why we modify the algorithm so that exact equivariance is lost.

The predictions in Algorithm 1 need not be of a uniform scale as they are the result of learned correlations with the group-equivariant layers. This could cause the scale of predictions affecting the alignment and capturing of the agreement in the creation of the deeper capsules. Thus, we introduce a layernorm that normalises all capsules across capsule-types and transformational states. This allows for Algorithm 1 to better follow the idea of importance and agreement that we discussed. This, however, causes a loss of exact equivariance. We justify this implementation by the following.

We see that the layernorm improves the performance of the models we train for classification, lending credence to our theory of the scale of predictions affecting the agreement among predictions. Moreover, many 'equivariant' models such as those in [1] and [2] have used techniques such as strided convolutions and layernorm to improve empirically observed performance at the expense of exact equivariance. We believe that the use of methods is justified as empirically-observed good model structure and mathematically-derived methods that build such structures are both important to build high performing models. The experimental results that we show in the next section justify our use of layernorm in Algorithm 1.

For all of the experiments, we use 32 16-dimensional capsule-types for the primary capsule layer and 3 capsule layers with 32 16-dimensional capsule-types for each of the layers. The class-capsule layer has as many capsule-types as the number of classes of the dataset in question. All the convolutional layers and prediction mechanisms are correlation operators defined over the p4 group [1].

The capsule layers for CIFAR10 and FashionMNIST use 2 iterations of Algorithm 1 and set k to 10. The capsule layers for CIFAR100 use 2 iterations and set k to 5. These hyperparameters were determined empirically.

**4.4 Training of our model**

The models were all trained using the cross-entropy loss and AdamW [20] with a OneCycleLR scheduler [21]. The models were trained for 150 epochs. The code was written in pytorch.

## 5. Results

The results of the experiments are given in Table 1, Table 2 and Table 3. Each table presents classification accuracies for the 5 test sets for a single dataset. The heading for each test datasets gives the extents of the translations and rotations in a tuple. Thus, the second column denotes test set 1, the third column denotes test 2 etc. We discuss the findings below.

First, we see that the task of transformed image-classification is not particularly easy. This is pronounced in the results on test dataset 5, where for almost all models test dataset 5 yields the lowest results. We also see that FashionMNIST is relatively easier as it has images with a uniform background and lower difficulty. CIFAR10 is harder to classify, and CIFAR100 is the hardest.

The capsule baselines CapsNet, EMCaps, and GCaps show a marked decrease in accuracies on CIFAR10 and CIFAR100, indicating their inability to learn diverse and complex objects.

We see that SOVNET and DeepCaps are stronger models, achieving better results. SOVNET shows better generalisation to transformations than the other baselines because of its equivariant modules. However, it (and DeepCaps) are not expressive enough to achieve the best results.

The residual models outperform the previously described baselines due to their architecture; particularly, P4MResNet18 and P4MResNet34 achieve much better generalisation and performance than the other models.

Our model, termed ICR, outperforms all the models on all the test sets, obtaining state-of-the-art results on transformed-image classification. This shows that equivariance and compositionality together with a carefully chosen architecture can improve the performance of a model. This is particularly important as ICR improves over the equivariant resnet models, something that many older capsule models were not able to do.

Table 1 The accuracies of various models on transformed classification for FashionMNIST. The training images have been translated by pixels up to 2 pixels and rotated by a random angle between (-180°, 180°). The results of the models on 5 test datasets have been given. **Our model achieves the best results on all the 5 test datasets.**

| Method | (0, 0°) | (2, 30°) | (0, 60°) | (2, 90°) | (2, 180°) |
|---|---|---|---|---|---|
| CapsNet | 86.90% | 84.94% | 84.93% | 84.75% | 84.72% |
| EMCaps | 82.99% | 82.67% | 82.18% | 82.32% | 82.18% |
| GCaps | 80.65% | 79.66% | 79.46% | 79.47% | 79.37% |
| DeepCaps | 92.07% | 91.71% | 91.70% | 91.76% | 91.66% |
| SOVNET | 94.11% | 93.77% | 93.56% | 93.57% | 93.60% |
| SOVNET-aug | 94.21% | 93.58% | 93.46% | 93.57% | 93.61% |
| ResNet-18-aug | 94.21% | 93.55% | 93.24% | 93.30% | 93.45% |
| ResNet-34-aug | 94.38% | 93.75% | 93.78% | 93.78% | 93.73% |
| P4MResNet-18-aug | 93.63% | 93.38% | 93.32% | 93.31% | 93.35% |
| P4MResNet-34-aug | 93.22% | 92.71% | 93.08% | 93.01% | 92.81% |
| **ICR (ours)** | **94.15%** | **94.13%** | **94.03%** | **94.09%** | **94.05%** |

Table 2 The accuracies of various models on transformed classification for CIFAR-10. The training images have been translated by pixels up to 2 pixels and rotated by a random angle between (-180°, 180°). The results of the models on 5 test datasets have been given. **Our model achieves the best results on all the 5 test datasets.**

| Method | (0, 0°) | (2, 30°) | (0, 60°) | (2, 90°) | (2, 180°) |
|---|---|---|---|---|---|
| CapsNet | 61.08% | 59.53% | 60.04% | 59.85% | 59.90% |
| EMCaps | 57.57% | 55.89% | 56.85% | 56.35% | 55.20% |
| GCaps | 39.09% | 41.03% | 41.43% | 41.25% | 41.08% |
| DeepCaps | 81.12% | 80.81% | 80.64% | 81.05% | 80.92% |
| SOVNET | 82.50% | 81.80% | 81.78% | 81.95% | 81.82% |
| SOVNET-aug | 80.14% | 79.64% | 79.94% | 79.99% | 79.65% |
| ResNet-18-aug | 78.84% | 79.28% | 79.72% | 79.60% | 78.95% |
| ResNet-34-aug | 81.27% | 81.15% | 81.44% | 81.60% | 81.65% |
| P4MResNet-18-aug | 89.88% | 89.46% | 89.33% | 89.54% | 89.41% |
| P4MResNet-34-aug | 89.12% | 89.02% | 89.18% | 88.85% | 89.10% |
| **ICR (ours)** | **92.08%** | **91.54%** | **91.24%** | **91.22%** | **91.48%** |

Table 3 The accuracies of various models on transformed classification for CIFAR100. The training images have been translated by pixels up to 2 pixels and rotated by a random angle between (-180°, 180°). The results of the models on 5 test datasets have been given. **Our model achieves the best results on all the 5 test datasets.**

| Method | (0, 0°) | (2, 30°) | (0, 60°) | (2, 90°) | (2, 180°) |
|---|---|---|---|---|---|
| SOVNET | 40.38% | 39.74% | 39.84% | 39.76% | 39.77% |
| SOVNET-aug | 40.38% | 39.82% | 39.69% | 39.78% | 39.99% |
| ResNet-18-aug | 50.03% | 50.56% | 51.15% | 51.00% | 51.14% |
| ResNet-34-aug | 51.40% | 51.86% | 51.49% | 51.93% | 52.11% |
| P4MResNet-18-aug | 64.22% | 64.19% | 63.86% | 63.89% | 63.38% |
| P4MResNet-34-aug | 66.12% | 65.90% | 65.66% | 65.56% | 65.92% |
| **ICR (ours)** | **69.10%** | **68.53%** | **68.41%** | **68.42%** | **68.13%** |

## 6. Conclusion

Our paper proposes a new routing algorithm for capsule networks. We term this algorithm 'iterative collaborative routing (ICR)'. ICR is an equivariant routing algorithm that allows for mathematical guarantees on transformation-robustness, and thereby guarantees for transformation-invariant learning of compositionality in visual objects [3].

We study the transformation-robustness of our proposed capsule network models with ICR, and see that it achieves state-of-the-art performance in classification under train and test geometric transformations.

Models that use ICR outperform a number of capsule, as well as residual, network baselines, showing that equivariance, compositionality, and a careful choice of model-architecture are important for high performance.

Our results can be seen as a step in the direction of showing that capsule networks can achieve high performance. Future work includes studying the role of compositionality and equivariance in more object-centric and diverse tasks such as visual question answering, where capsules can correspond directly to objects, and routing can correspond directly to seen relationships.